# Diffuse Map Guiding Unsupervised Generative Adversarial Network for SVBRDF Estimation

1st Zhiyao Luo
*Computer Science and Technology*
*Jinan University*
Guangzhou, China
zhiyaoluo.leo@gmail.com

2st Hongnan Chen
*State Key Lab for Novel Software Technology*
*Nanjing University*
Nanjing, China
MG21330010@smail.nju.edu.cn

*Abstract*—Reconstructing materials in the real world has always been a difficult problem in computer graphics. Accurately reconstructing the material in the real world is critical in the field of realistic rendering. Traditionally, materials in computer graphics are mapped by an artist, then mapped onto a geometric model by coordinate transformation, and finally rendered with a rendering engine to get realistic materials. For opaque objects, the industry commonly uses physical-based bidirectional reflectance distribution function (BRDF) rendering models for material modeling. The commonly used physical-based rendering models are Cook-Torrance BRDF, Disney BRDF. In this paper, we use the Cook-Torrance model to reconstruct the materials. The SVBRDF material parameters include Normal, Diffuse, Specular and Roughness. This paper presents a Diffuse map guiding material estimation method based on the Generative Adversarial Network(GAN). This method can predict plausible SVBRDF maps with global features using only a few pictures taken by the mobile phone. The main contributions of this paper are: 1) We preprocess a small number of input pictures to produce a large number of non-repeating pictures for training to reduce over-fitting. 2) We use a novel method to directly obtain the guessed diffuse map with global characteristics, which provides more prior information for the training process. 3) We improve the network architecture of the generator so that it can generate fine details of normal maps and reduce the possibility to generate over-flat normal maps. The method used in this paper can obtain prior knowledge without using dataset training, which greatly reduces the difficulty of material reconstruction and saves a lot of time to generate and calibrate datasets.

## I. INTRODUCTION

In the development of computer graphics, substantial attention has been paid to produce awesome visual effects and render a virtual world as fine as the real world. Animation studios and game companies use these techniques to create realistic graphics that give audiences an immersive visual effect. To make this goal possible, researchers make effort on proposing more complex shading model, which can better describe the material and fit the real world lighting effect. Moreover, they cooperate with the artists to create and paint textures, which reflect sophisticated appearance in a simple way.

Material acquisition has been spotlighted in the past ten years due to the booming development of machine learning, which make it possible to extract real-world materials from images. Machine learning based material extraction changes the situations that artists should spend a long time to paint the textures and researchers should use prohibitively expensive devices to reconstruct materials. Despite these considerable advantages, the precision of the estimated appearance is still a long-standing problem due to the high dimensionality of appearance data. Material is the properties of an object that determine how that object interacts with light in the environment. To reduce the sophistication of material properties reconstruction, Bi-directional Reflectance Distribution Function(BRDF), a ratio of reflected radiance to incident irradiance, is used to defines how light is reflected at an opaque surface. As can be seen, the high dimensionality of appearance data is from an 6D spatially-varying bidirectional reflectance distribution function (SVBRDF) and it can only express the surface reflectance of opaque materials. It is interesting to note that the lack of light information has made it challenging to reconstruct SVBRDF maps. The inherent ambiguity is a more intricate issue, since we should provide a strong prior to solve the under-constrained problem and disentangle SVBRDF maps. In an effort to overcome these challenges, considerable researches have been conducted to create a large set of SVBRDF maps data and train a powerful prior supervised by the dataset. Nevertheless, each SVBRDF map in the large scale dataset is time-consuming for artists to paint, which is nontrivial to obtain.

In recent years, Generative Adversarial Network(GAN) (Goodfellow et al., 2014) based SVBRDF reconstruction method has been adopted to alleviate the reliance on a large volumes of data. To be more specific, it enables SVBRDF estimation without initialisations and datasets. It is interesting to note that the guessed diffuse map plays a crucial role in guiding the network to generate plausible SVBRDF maps. Nonetheless, this success is based on a hypothesis that the material acquired is stationary, namely the features of the material duplicate themselves periodically. The failure of global features reconstruction is caused by the poor method to obtain reasonable guessed diffuse maps. Therefore, SVBRDF maps with global features reconstructions are still facing challenges and it is worthwhile devoting much effort to this.

In this work, I introduce a diffuse map guiding method for the first time to estimate plausible SVBRDF maps with global features. In general, supervised learning has been widely used to estimate SVBRDF maps through providing a strong prior, which is based on a large scale datasets with thousands of artist-created SVBRDF maps. However, creating these maps is time-consuming and costly. Therefore, this method will be hindered by a lack of precise datasets. Impressively, the method of reflective properties reconstructions proposed in this study fundamentally solves the problem. It might pave a new way to lightweight material reconstructions and open the door to research on appearance estimation without initialisations and datasets.

## II. RELATED WORK

Substantial studies have been undertaken in SVBRDF estimation fields in decades. Most of them are focus on supervised learning methods. Unsupervised learning methods have only recently been applied to SVBRDF extraction and have many aspects of limitations. In the following I will give a chronological overview of the important work in this area.

The material in the real world is very complex. For opaque objects, it includes optical phenomena such as reflection and scattering, so it is very difficult to reconstruct the SVBRDF of a material. Traditionally, these materials are collected by a prohibitive machine, namely spherical gantry. It samples the material by enumerating the direction of the light source and the direction of the camera, and then measures the incident radiance, which results in a large amount of data. To reduce the cost of sampling, researchers are working to collect data using lightweight devices. (Aittala et al., 2015) assume that the texture-like is self-similarity, that is, points on the material have similar reflectance properties. Using this self-similarity, they restored the reflectance properties of the material from photos taken by two mobile phones (one with a flash and one without) with optimization. (Aittala et al., 2016) proposed a method that combines inverse rendering with convolution neural network (CNN) to extract a spatially varying parametric reflectance models from a flash-illuminated photo. However, these method is only applicable to stationary materials, which means that the material beyond the assumptions cannot be successfully reconstructed.

Recently, deep learning-based methods have made significant progress in SVBRDF estimation for a single image. (Xiao et al., 2017) used self-augment training strategies to train the convoluted nerve network. It only requires a small number of labeled SVBRDF datasets to learn a large number of photos of unlabeled spatially varying materials. (Wenjie et al., 2018) improved this method by using only the unlabeled training dataset to complete the training. (Deschaintre et al., 2018) also used convolution neural networks to estimate reflectance characteristics from a single photo illuminated by a flash. Unlike before, they added global features extracted from each phase of the U-net architecture. This method takes into account the global information provided by local lighting to enrich texture details. In addition, they introduced a rendering loss to enhance the estimated reflectance parameters by rendering the estimated SBVRDF map and comparing it with the input image. (Deschaintre et al., 2019) proposed a learning-based SVBRDF estimation system that supports any number of inputs. They first use a network (similar to (Deschaintre et al., 2018)) to extract latent features from each individual input image. These features are then combined and sent to an order in dependent pooling layer to fuse multiple feature maps from a single image network to generate the final SVBRDFs. This method helps to extract visual cues scattered in different inputs and combine them, adding latent features to the resulting SVBRDF maps. The estimated SVBRDF maps have more detail than the single-image based extraction method. (Gao et al., 2019) proposed a method based on the combination of reverse rendering and optimization, and obtained a priori by learning a large amount of data to reconstruct SVBRDFs from any number of inputs. They train a full-convolution auto-encoder and then optimize it in this space. This ensures the rationality of the rebuilt SVBRDFs. However, this method requires a good initial value for SVBRDF maps, such as the results from a single image method (Deschaintre et al., 2018; Li et al., 2018). Once they get their initial values, they code them and optimize their latent space. As the number of input images increases, the SVBRDF maps become more accurate, but still largely depend on the quality of the initial values. (Guo Yu et al., 2020) proposed MaterialGAN, which is a deep convolution neural network based on StyleGAN2 (Karras et al., 2019). It generates reasonable materials from a large, spatially varying dataset of materials (Deschaintre et al., 2018). They optimize the potential representation of the MaterialGAN to produce a material map that matches the appearance of the image captured during rendering. (Guo Jie et al., 2021) proposed a two-stream network as a new variant of standard (ST) convolution. They extract features from two different branches of HA and ST convolutions, respectively. They use HA convolution to guess saturated pixels from the surrounding unsaturated areas, thereby reducing the effect of specular highlights on diffuse maps. This works well for images with strong highlights and avoids the contamination of the images by highlights. This method only takes a single picture as input and does not require camera and light source parameters to output high quality material parameters. (Zhao et al., 2020) proposed an unsupervised recovery method for SVBRDF maps. They trained SVBRDF maps using a GAN architecture and calculated resistance losses. To make the results more realistic, they use guessed diffuse maps as a criterion to provide more information. The joint loss includes adversarial loss and diffuse loss. Their network can predict a reasonable SVBRDF map from a single input image and does not require any datasets. They use an encoder-decoder architecture to provide high-quality texture synthesis. However, when the input image has strong highlights, this method is difficult to solve the high-light pollution caused by saturated pixels. Vecchio et al.(2021) purpose a semi-supervise GAN based method for SVBRDF estimations and obtain state-of-the-art performance synthetic images dataset.

## III. DIFFUSE MAP GUIDING GAN FOR SVBRDF ESTIMATION

This chapter focuses on a SVBRDF estimation method without dataset. In practice, SVBRDF data comes from two sources. One is direct measurements of real objects by researchers or large companies using prohibitive machine. Another is directly mapped by the artists. In recent years, material reconstruction methods which are based on machine learning have been proposed, but most existing methods rely on a very large SVBRDF datasets. (Zhao et al., 2020) proposed a new unsupervised generative adversarial network to handle this problem. Nevertheless, this method has the limitation of not being able to estimate the materials with global structure. In this paper, we purpose a novel method which can reconstruct the SVBRDF maps with global structures without a given training dataset. In this chapter, I will firstly introduce the overall pipeline in brief, and then elaborate the following parts in detail, which includes data acquisition, the process of extracting diffuse maps, network architecture, renderer architecture, loss functions and trainning strategy.

### A. SVBRDF estimation pipeline

Before introducing our pipeline, we need to review the work of (Zhao et al., 2020). This is because the work of (Zhao et al., 2020) provides a reference for our pipeline. (Zhao et al., 2020) proposed a generative adversarial network for SVBRDF recovery and synthesis. This network system is unsupervised, so no datasets are required for training. It's inside the entire pipeline, and they firstly input a high-resolution stationary image that was taken. Thereafter, different pictures which are clipped randomly from the input images, are used to train the network. They input the clipped pictures into an untrained generator consisting of one encoder and two decoders. The generator outputs SVBRDF maps, then the renderer renders the SVBRDF maps, and finally utilise a discriminator to compare the render result with the ground true picture.

In our pipeline, we utilise a set of multi-view images, which are scanned by a mobile phone as inputs. For the training steps, the number of pictures taken manually is not enough. Unlike previous work, in order to increase the amount of training data and reduce the over-fitting, we first preprocess the input pictures. The detail of preprocessing process will be mentioned in the following sections. Then, we used a new method to extract the plausible diffuse map from these multi-view images. This method is a good way to extract diffuse maps of materials with global features. Afterwards, we input the preprocessed pictures into the generator whose encoder-decoder structure has been improved. Next, the generator will output the estimated SVBRDF maps. In order to provide more prior, the generated diffuse map is compared with the extracted diffuse map. Finally, after rendering the map using the rendering module, the rendering results are compared with the ground true image with the discriminator. The adversarial loss and the diffuse map loss will guide the generator to produce SVBRDF maps more plausible in the following iterations.

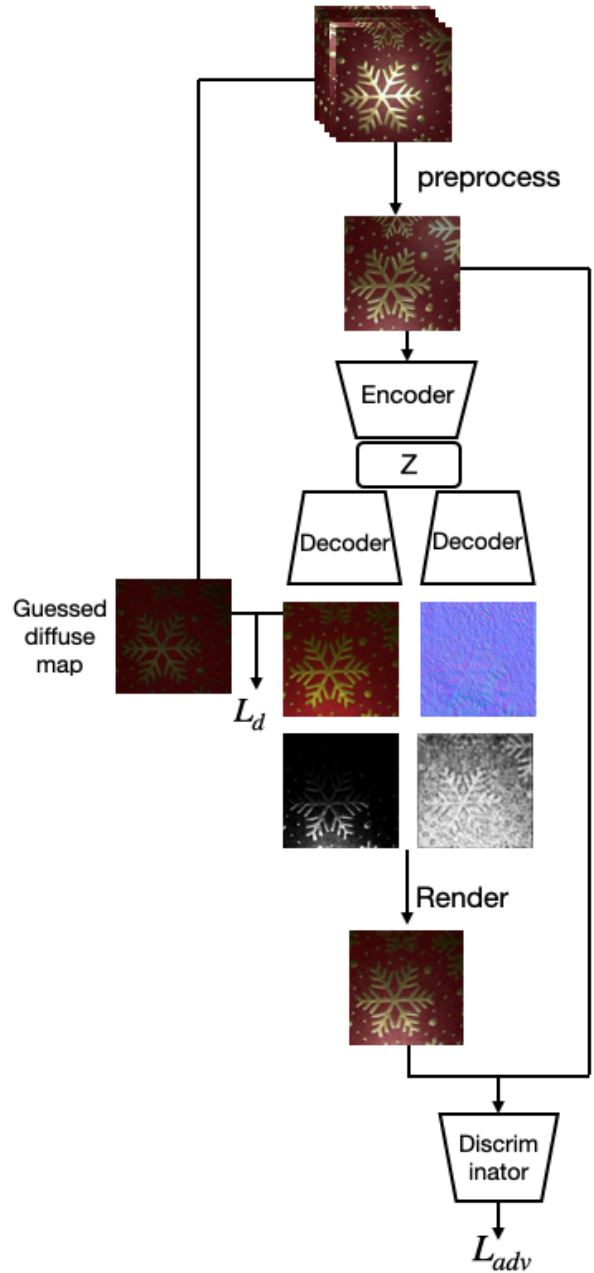

*Figure1: Our pipeline*

### B. Data acquisition and preprocessing

In the real-world material data acquisition step, we assume that the light source is very close to the camera. Moreover, in order to get an image without distortion, we should place the phone to be parallel to the material surface. With the flash on, we take several pictures at the same height and different positions. Therefore, for each individual pixel, the illumination direction can be considered the same as the viewing direction. To make the result comparison step more easier, we chose to utilise (Guo Yu et al. 2020) datasets directly in our experiment. In their dataset, each material set has nine pictures taken from different positions. Then they project these images into a common frontal view. This dataset contains a wide variety of materials, such as wood, greeting cards, plastics, marble, and so on, covering most of the isotropic materials. Therefore, it can greatly test the performance of our method.

## C. Diffuse maps estimation

Zhao et al.(2020) proposed a novel diffuse map estimation method based on the process of computation as in (Aittala et al., 2016). They use the estimated diffuse map as the ground truth, called guessed diffuse map. This is because their method is unsupervised and they do not have the ground truth diffuse map. Therefore, they have to utilise the guessed diffuse map to provide a stronger prior. However, they only manipulate the images with a simple averaging and Gaussian blurring to obtain a guessed diffuse map. So it is only applicable for stationary material without saturated pixel pollution caused by highlights. In our experiment, this method will blur textures with global features. Therefore, for materials with global features, we cannot use this method to get a plausible guessed diffuse map. The failure to estimate a diffuse map with global features is also one of the reasons why material reconstructions with this method cannot have global features. In addition, we observed experimentally that diffuse maps obtained by this method still contain information of normal maps and specular maps, which is detrimental to recover the disentangled material maps.

To solve this problem, we utilised a novel method to obtain guessed diffuse map with global features. Before introducing our method, we need to review the work of multi-view diffuse texture extraction (Peter and Hannes 2015). They first used mobile phones to take pictures of the material surface. In this step, they need to use internal measurement unit (IMU) which includes gyroscope, accelerator and magnetometer. Then select a group of pictures as shown in Figure 2. Next, they utilise SIFT features (Lowe D., 1999) to detect the image features and calculate the homology to align the reference (the first) image. Finally, for each pixel of reference image, find the minimum value of the corresponding pixel of each image. The result is diffuse map.

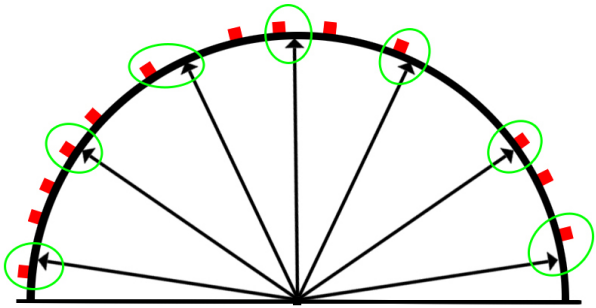

*Figure2: Select the direction of input images (Peter and Hannes, 2015)*

Note: The red squares are the position we capture images, and the arrows are the directions we want to sample on the hemisphere. We choose the red squares closest to arrows, which are circled in green.

In the experiment, they (Peter and Hannes, 2015) observed that specific reflection only increases the reflected radiation, which is different from pure reflection. Therefore, they assume that material reflection is composed of diffuse and specular. Therefore, for rendering equation, BRDF items can be split into diffuse and specific terms.

In Cook-Torrance BRDF model, the diffuse reflection term is Lambertian, which means that reflected diffuse light $\rho_d(x) \cdot E(x)$ is the same for all directions on the hemisphere above the surface. Therefore, the radiance of every pixel in image may consist of only diffuse light $k_d \cdot f_{lambert}$ or may be combined with specular reflection $k_s \cdot f_{cook-torrance}$. In this scenario, we can conclude that for each pixel in images, we just need to find out the corresponding minimum radiance pixel amid captured images. Then, all of these minimum radiance pixels can make up an estimated diffuse map with global features.

Nevertheless, it is almost impossible for us to capture images with internal measurement unit (IMU). This is because it is such a probative machine which includes gyroscope, accelerator and magnetometer. Therefore, in our experience, the images are just captured in an easy way as mention above. Albeit this simple method of image acquisition does not take into account the effect of radiation attenuation, the estimated diffuse map is also very close to reality.

## D. Network architecture

Our work is based on a generative adversarial network which possess a two-stream generator and a discriminator. The design of two-stream generator is based on the work purposed by Zhao et al.(2020). Our generator uses the auto-encoder structure. It consists of two parts: an encoder and a decoder. Encoder is used to encode the input data, extract high-dimensional features, map to the low dimensional latent space, and get the encoded data. The decoder is used to restore the encoded data in the latent space as much as possible as the input data. The restored data has some information loss compared to the initial input data, so it is not really the input data. Our generator includes an encoder $E_n$ and two decoders $D_{n,r}$ and $D_{\rho_d \rho_s}$. While $D_{n,r}$ generates normal and roughness maps, $D_{\rho_d \rho_s}$ generates diffuse and specular maps. In our experiment, we did not pre-train the network, that is, encoder is untrained. Fortunately, according to output of several layers visualisation by the encoder, the features of textures are preserved well. (Zhao et al., 2020) Therefore, we also use the simple untrained encoder structure similar to (Zhao et al., 2020).

However, in our experiments, we found that the normal map is over-flat due to the weak ability of normal map extraction in this network architecture. After analysis, we believe that it is caused by the lack of ability for encoder to extract features from normal map. This results in rendered SVBRDF maps are very different from the real material. Therefore, we have improved the structure of auto-encoder.

In our network, the encoder has 5 convolutional layers, with each followed by an instance normalisation and leaky-ReLu activation. As you can see in Figure 3-3 , We include more layers in the encoder design, which helps the encoder extract more detailed features. After the latent space with high-dimensional features is obtained through encoder, two decoders are designed to reconstruct the material information. As can been seen in Figure , $D_{n,r}$ decoder has a convolution layer and three deconvolution layers, the first three being closely followed by an instance norm layer and Leaky ReLU activation layer. The last layer uses the Tanh activation layer. $D_{p_d,p_s}$ decoder has only three deconvolution layers, the first two being followed by an instance norm layer and Leaky ReLU activation layer, and the last using the Tanh activation layer. For $D_{n,r}$ decoder, we use fewer convolution layers, since we need to avoid blurring the diffuse map. The reason we decode $D_{n,r}$ and $D_{p_d,p_s}$ separately is that previous studies have shown that using only one decoder causes decoder mistaking features of normal as features of diffuse because decoder lacks ability to disentangle diffuse map and normal map.

We use Discriminator to compare the rendered results of estimated SVBRDF maps with input image to continuously optimize the Generator to generate the estimated SVBRDF maps that are closer to the input image after rendering. As shown in the Figure n, the Discriminator consists of five convolution layers. The first four convolution layers are followed by instance norm layer and Leaky ReLU activation layer, and the last one uses Sigmoid activation layer.

*E. Network architecture*

The renderer architecture includes light source, camera and material. We set the light source in the same location as the camera, which is the same as Zhao et al. (2020). So for each pixel, the illumination direction is the same as the observation direction. We calculate the BRDF results on the micro-facet model with the material parameters on estimated SVBRDF maps. To simplify the calculation, we have not considered the model of light attenuation in propagation.

*F. Training strategy*

In our experiment, there are two possibilities for network input: one is to randomly select one original image from multiple pictures for input, the other is to randomly select two pictures and blend them together with random weights. The two possibilities are equal. After entering the picture into the Generator, the SVBRDF maps output by the generator are rendered with a renderer, and the resulting image is put into Discriminator to compare and discriminate with the original image.

At the beginning of our experiment, we tried to use 20,000 iterations. Each iteration uses a 5-step gradient descent algorithm for $D_{n,r}$, a 1-step gradient descent algorithm for $D_{p_d,p_s}$, and a $2 \cdot e^{-5}$ learning rate. However, this method may lead to over-fitting for a situation where there is only a small amount of training data. In addition, this training method takes a long time. We use P100 GPU to train the network, which takes about 3 hours for one material.

To solve the fitting problem and reduce the training time, we use two-stage training strategy (Wen et al., 2022). We first used a picture with global features for 10,000 iterations. For the first training, we do not need the result of SVBRDF maps, we only need to save this model parameter after 10,000 iterations of training. This pre-trained model can be reused. Subsequently, we input pictures which we require the estimated SVBRDF maps and continued 15,000 iterations using pre-training model parameters. The results obtained are more plausible than those obtained by one-stage training methods, effectively alleviating the over-fitting phenomenon. We also used P100 GPU for training. In this scenario, it takes about 2 hours in the second step of 15,000 iterations training, which effectively shortens the training time. The reason this training method can reduce overfitting and training time is that pre-training provides the generator with a prior, or a reasonable initialisation, including normal map with color close to blue, input image with color mostly determining the color of diffuse map, rough map with color close to gray, specular map with darker color. With these prior, generator can generate almost plausible SVBRDF maps faster, which means it requires fewer iterations. Fewer iterations reduce the probability of inputting a large number of identical pictures. This is because each iteration we need input new pictures. These input images are generated by a random combination of a few captured photos. The more pictures are generated, the greater the probability of duplicate pictures, which will lead to overfitting.

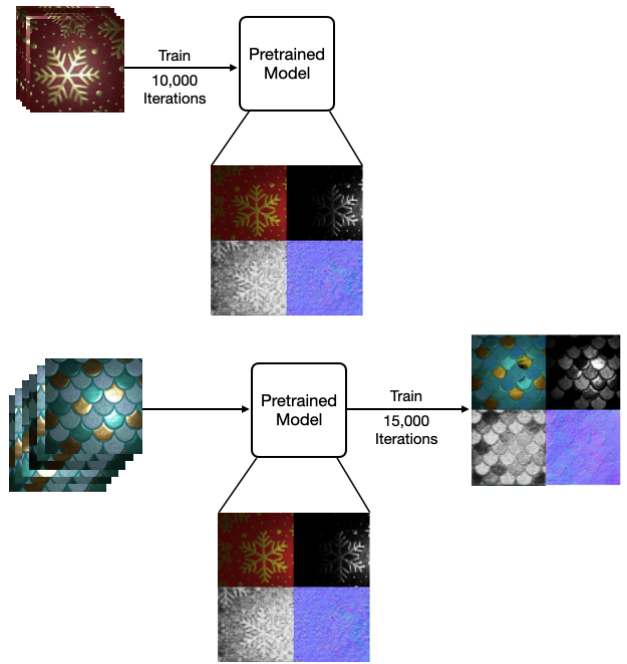

*Figure3: Select the direction of input images (Peter and Hannes, 2015)*

## IV. EXPERIMENTAL RESULT

This chapter describes the reconstruction result of our method. To facilitate comparisons with related results, we selected the (Guo Jie et al. 2020) dataset. All photos in this dataset are taken with the flash turned on by the mobile phone. For each material, the dataset provides nine pictures taken in different positions. These pictures reflect the appearance of real-world materials. We will then compare and analyze the results of related work. Finally, we will point out the limitations of our approach and point out the direction for future improvement.

### A. Comparison with prior work on real data

Albeit our method only use nine images as training data, it seems achieves better performance than those training with thousands of images datasets like Deschaintre et al.(2018), Deschaintre et al.(2019), and Gao et al.(2019). Gao et al. (2019) must rely on reliable initialization parameters to optimize for plausible results, and we can rely on no initialization parameters. As can be seen in set 1, Gao et al. (2019) gives a wrong estimated SVBRDF maps which the estimated diffuse map is black, while our work still can give a reasonable SVBRDF maps which diffuse map can basically represent the texture and base color of the material. Despite some mismatches due to the ambiguous highlight spot and the flat normal map due to the ambiguity of diffuse map and normal map, our method still recovers major features that are close to those generated by MaterialGAN (Guo Yu et al.,2020) and Highlight-Aware Two-Stream Network (Guo Jie et al.,2021). Moreover, we can observe that, thanks to the accurate guessed diffuse map guiding, our estimated diffuse map is more clear, sharper and less prone to overfitting (burn-in) than Guo Jie et al.(2021) in set 2. Our reconstructions result can also achieve plausible and realistic, comparing with other methods which are based on a large scale of datasets. We can observe that the global features global features can be well reconstructed in SVBRDF maps.

However, our method is affected by the contamination of saturated pixels caused by highlights because our method cannot handle ambiguous highlightspot. Our specular map in set 1 and set 3 is a good reflection of this phenomenon and our diffuse map in set 1 and set 3 Actually, it is a recognized challenge in the field of material reconstructions. We can find this problem in others works like Gao et al.(2019) in set 1 and MaterialGAN (Guo Yu et al.,2020) in set 3. According to our analysis, for pictures with strong highlights, our method will learn highlights as features, which will result in the artifact of the maps and unable to recover the plausible estimated SVBRDF maps.

Input      SVBRDF      Render

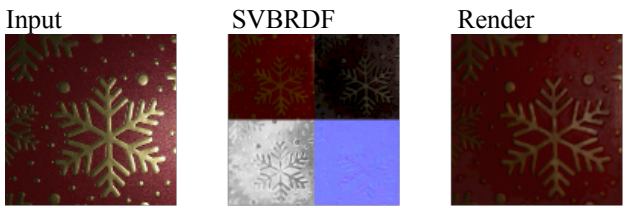

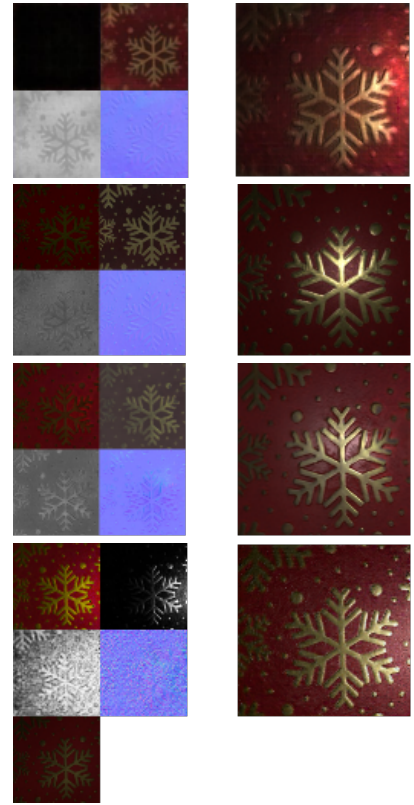

*Figure4: Result set 1*

Note: The first column is the input image. The second column is the corresponding estimated SVBRDF maps, which includes Diffuse (top-left), Specular (top-right), Roughness (bottom-left), Normal (bottom-right) and guessed diffuse map (only in Ours). The first row is the result of Deschaintre et al. (2018), the second row is the result of Gao et al.(2019), the third row is the result of Guo Yu et al.(2020), the forth row is the result of Guo Jie et al(2021) and the fifth row is our result.

Input      SVBRDF      Render

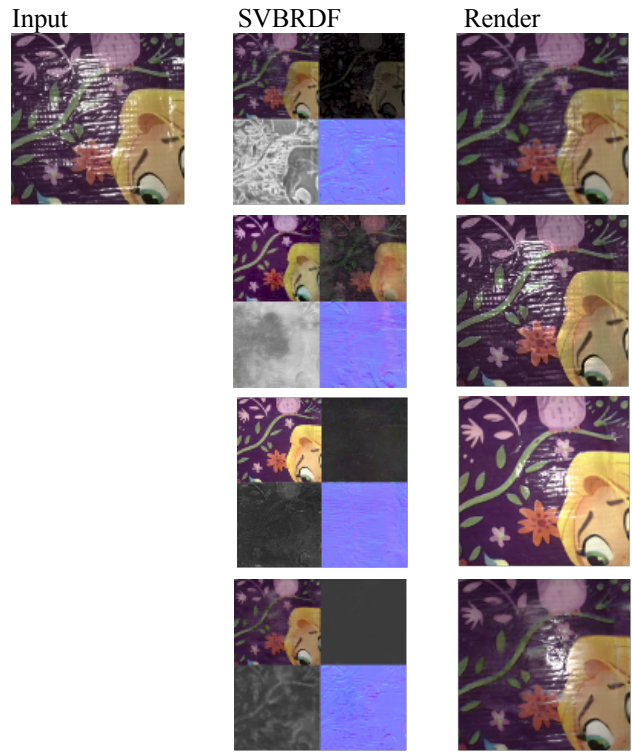

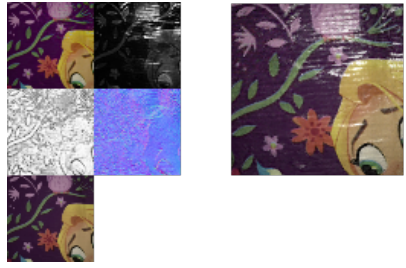

*Figure5: Result set 2*

*Note: The first column is the input image. The second column is the corresponding estimated SVBRDF maps, which includes Diffuse (top-left), Specular (top-right), Roughness (bottom-left), Normal (bottom-right) and guessed diffuse map (only in Ours). The first row is the result of Deschaintre et al. (2018), the second row is the result of Gao et al.(2019), the third row is the result of Guo Yu et al.(2020), the forth row is the result of Guo Jie et al(2021) and the fifth row is our result.*

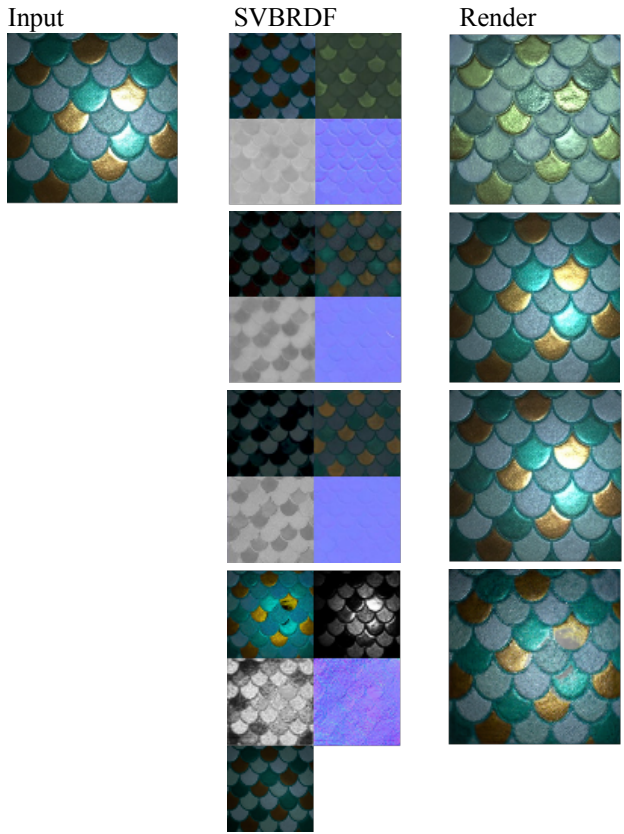

*Figure6: Result set 3*

*Note: The first column is the input image. The second column is the corresponding estimated SVBRDF maps, which includes Diffuse (top-left), Specular (top-right), Roughness (bottom-left), Normal (bottom-right) and guessed diffuse map (only in Ours). The first row is the result of Deschaintre et al. (2019), the second row is the result of Gao et al.(2019), the third row is the result of Guo Yu et al.(2020) and the forth row is our result.*

### B. Comparison with prior work on synthetic data

It is clear that our approach does not perform as well as real data in the synthetic data in both images result and RMSE. This is because thousands of training data provides a prior to supervised learning methods, enabling them to generate ideal smooth normal maps. Without this prior, our normal map is bumpy and very different from the synthetic ground true SVBRDF maps. However, this ideal smooth normal map does not exist in real life material. So our approach fits real data better, so it works better in real data. Moreover, our specular map uses only one channel, which means it can only represent the intensity of the specular reflectance but not the color of the specular reflectance. This makes it impossible to reconstruct some materials with special reflective properties. For example, the highlight color in the set 5 is brown, and our specular map cannot represent the highlight color, so the highlight feature is mapped onto the diffuse map, causing the contamination of diffuse map. Therefore, we need to expand the specular map from a one channel to RGB channels to accommodate more materials with complex specular reflectance. To reduce the high-light pollution in diffuse map, we can choose guessed diffuse map instead of estimated diffuse map to render and see if the result is improved. For example, in set 5, we used guessed diffuse map to render, alleviating the pollution caused by the brown specular reflectance. As you can see in our result, albeit specular property is not well reconstructed, plausible results are still available. It is worth noting that although our normal map has a larger RMSE value in set 4 due to the bump of our normal map, our normal map will have better performance in terms of material representation. This is because the over-flat problem in the normal maps of Gao et al. (2019) and Guo Yu et al. (2020) makes the material insensitive to changes in light. According to our analysis, this is because normal map and diffuse map cannot completely disentangle, resulting in the feature of normal map being mapped to diffuse map, that is, the shadow on diffuse map should be the feature of normal map. The shadow caused by self-occlusion is a recognized challenge in the field of material reconstructions, and our approach has the same problem.

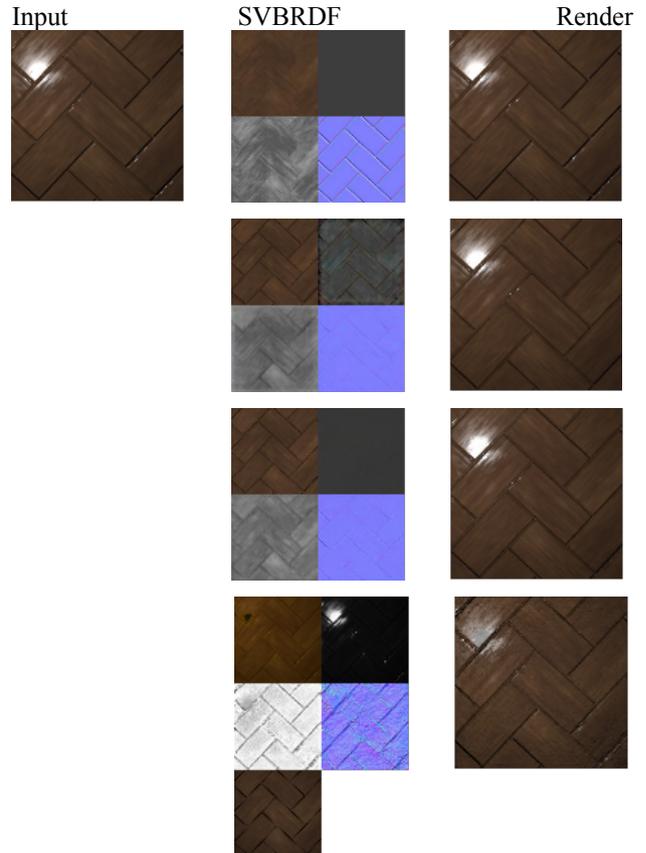

*Figure6: Result set 4*

*Note: The first column is the input image. The second column is the corresponding estimated SVBRDF maps, which includes Diffuse (top-left), Specular (top-right), Roughness (bottom-left), Normal (bottom-right) and guessed diffuse map*

*(only in Ours). The first row is the ground-truth, the second row is the result of Gao et al.(2019), the third row is the result of Guo Yu et al.(2020) and the forth row is our result.*

| Method | Diffuse | Specular | Roughness | Normal | Guessed Diffuse |
|---|---|---|---|---|---|
| Gao et al. (2019) | 0.031 | 0.054 | 0.059 | 0.067 | |
| Guo Yu et al. (2020) | 0.031 | 0.024 | 0.047 | 0.068 | |
| Ours | 0.085 | 0.208 | 0.451 | 0.143 | 0.071 |

*Table 1: RMSE of result set 4*

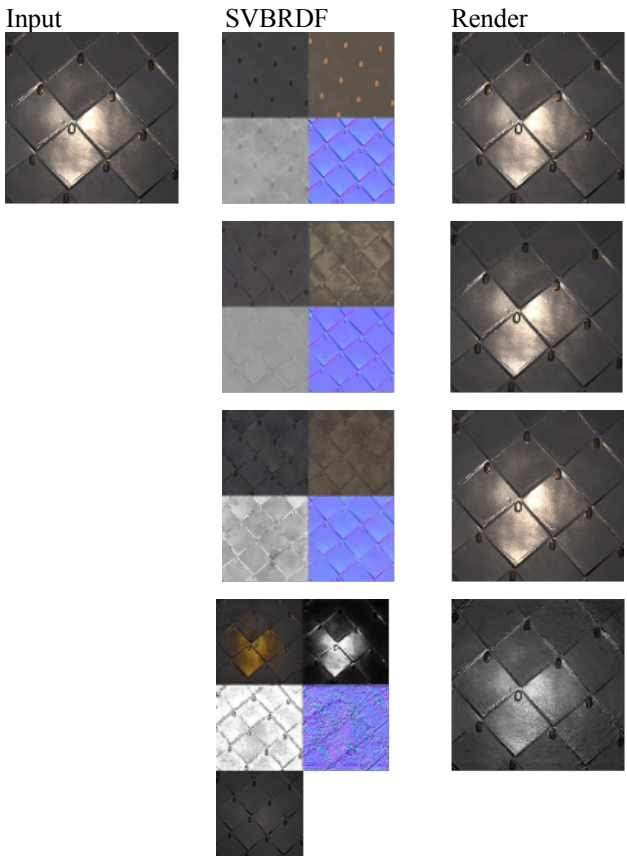

*Figure6: Result set 5*

*Note: The first column is the input image. The second column is the corresponding estimated SVBRDF maps, which includes Diffuse (top-left), Specular (top-right), Roughness (bottom-left), Normal (bottom-right) and guessed diffuse map (only in Ours). The first row is the ground-truth, the second row is the result of Gao et al.(2019), the third row is the result of Guo Yu et al.(2020) and the forth row is our result.*

| Method | Diffuse | Specular | Roughness | Normal | Guessed Diffuse |
|---|---|---|---|---|---|
| Gao et al. (2019) | 0.029 | 0.071 | 0.035 | 0.039 | |
| Guo Yu et al. (2020) | 0.035 | 0.058 | 0.102 | 0.047 | |
| Ours | 0.089 | 0.277 | 0.284 | 0.157 | 0.070 |

*Table 2: RMSE of result set 5*

### C. Limitations And Further Improvement

In addition to the saturated pixel contamination caused by the highlights, the over-flat problem of normal map, the shadow problem caused by self-occlusion, and the problem of unable to express specular reflectance color we mentioned earlier, our method also has the problem of over-reliance on input images, that is, using different pictures in the same material dataset for input will result in different SVBRDF results. In addition, it is difficult to estimate the appropriate SVBRDF maps for the darker parts of the picture. For further improvements, we will introduce highlight-aware (HA) convolution operation (Guo Jie et al., 2021) to handle high-light contamination problem. Moreover, we will try to use the analysis model (Mashhuda et al. 2008) to guess the reasonable normal map and use it to provide more prior, which may solve the over-flat problem and the shadow problem caused by self-occlusion. To address the impact of input images on the output, we will attempt to use Variational Autoencoder (VAE) as the generator, which will ease the dependence of the results on the input images.

### IV. CONCLUSION

Our approach uses unsupervised learning to avoid reliance on datasets, which reduces time costs. Compared with previous work, we can reconstruct materials with global features without datasets. Our main work includes four parts. First we preprocess the dataset to alleviate the problem of over-fitting caused by insufficient training data. In addition, we introduce a novel method to generate guessed diffuse map, which is one of the keys to the success. Since this new method can generate a clear diffuse map with global features, whereas the guessed diffuse map generated by the previous method will be very fuzzy. Zhao et al. (2020) suffer from over-flat normal map, so we improved the structure of auto-encoder to get a clearer and sharper normal map. Finally, we used two-stage training strategy (Wen et al. 2022) to train our network, and we observed that the over-fitting phenomenon would be further alleviated. Our method can generate plausible SVBRDF maps for real life materials. Albeit our results still suffer from common problems in material reconstructions such as high light pollution, self-occlusion shadow, over-flat normal map, and inherent ambiguity, the results are comparable to supervised learning methods which are using a large number of datasets. Our results can produce plausible result when they are used to render in practical and achieve the expected results.